\pdfoutput=1
\documentclass[10pt,twocolumn,letterpaper]{article}

\usepackage{btas}
\usepackage{times}
\usepackage{epsfig}
\usepackage{graphicx}
\usepackage{amsmath}
\usepackage{amssymb}
\usepackage{url}
\usepackage{multirow}
\usepackage{booktabs}
\usepackage{tabu}

\btasfinalcopy 

\ifbtasfinal\fi

\newcommand{\tabincell}[2]{\begin{tabular}{@{}#1@{}}#2\end{tabular}} 

\begin{document}
	
	\title{ICFVR~2017: 3rd International Competition on Finger Vein Recognition}
	
	\author{$^*$Yi Zhang$^1$, $^*$Houjun Huang$^1$, Haifeng Zhang$^1$, Liao Ni$^1$, \\
		 Wei Xu$^2$, Nasir Uddin Ahmed$^3$, Md. Shakil Ahmed$^4$, Yilun Jin$^1$, \\ 
		 Yingjie Chen$^1$, Jingxuan Wen$^1$ and Wenxin Li$^1$\\
		$^1$Peking University\\
		$^2$Shenzhen Maidi Technology Co., LTD.\\
		$^3$TigerIT\\
		$^4$Chittagong University of Engineering \& Technology \\
		$^*$ These authors contributed equally to this work 
	}
	
	\maketitle
	\thispagestyle{empty}
	
	\begin{abstract}
		In recent years, finger vein recognition has become an important sub-field in biometrics and been applied to real-world applications. The development of finger vein recognition algorithms heavily depends on large-scale real-world data sets. In order to motivate research on finger vein recognition, we released the largest finger vein data set up to now and hold finger vein recognition competitions based on our data set every year. In 2017, International Competition on Finger Vein Recognition (ICFVR) is held jointly with IJCB 2017. 11 teams registered and 10 of them joined the final evaluation. The winner of this year dramatically improved the EER from 2.64\% to 0.483\% compared to the winner of last year. In this paper, we introduce the process and results of ICFVR 2017 and give insights on development of state-of-art finger vein recognition algorithms.
	\end{abstract}
	
    \section{Acknowledgements}
    
    This paper is partially supported by the National Natural Science Foundation of China (NSFC Grant Nos. 91646202 and 61472006) and the Seeding Grant for Medicine and Information Sciences of Peking University.
    
	\section{Introduction}
	
	Finger vein recognition has become a new method of personal identification because of its non-contact, live sampling, high accuracy and fast authentication features. With the development of finger vein recognition technology, it draws more and more attention of academic community and has been successfully applied to real-world applications.
	
	Like other sub-fields of biometrics, data sets play an important role in developing recognition algorithms for finger vein. However, finger vein community faces some difficulties in data sets. Firstly, there are rarely available data sets for researchers. To our best knowledge, only a few universities or organisations provide public data sets \cite{lu2013available,Tome_IEEEBIOSIG2014,Tome_ICB2015_AntiSpoofFVCompetition}. Secondly, the public data sets contain relatively small quantities of finger images. The largest one (except ours) is \cite{kumar2012human} which provides images of about three hundred fingers. This quantity is significantly smaller than those in other biometrics sub-fields like face recognition ~\cite{yi2014learning,wolf2011face} and fingerprint identification ~\cite{fingerprint-web}. Thirdly, there is not currently available a professional online platform for researchers to process large-scale finger vein data sets and evaluate their algorithms before our platform published.  To help the community dealing with these problems, we have collected a finger vein data set from large-scale real-world applications and provided an online platform (RATE) for researchers to evaluate their algorithms on our released data sets. For now, researchers can access images of fifty fingers and can run their algorithms on two data sets including 2,000 fingers on our online platform. 
	
	
	In recent years, new machine learning algorithms appeared and some of them have been introduced to the field of biometrics successfully. In order to encourage development of new algorithms for finger vein recognition, we held FVRC~2015~\cite{xian2015icb} and FVRC~2016~\cite{ye2016fvrc2016} competitions, jointly with ICB 2015 and 2016 respectively. This year, we continued to hold International Competition on Finger Vein Recognition (ICFVR) 2017 jointly with IJCB 2017. We also welcomed researchers to adapt their finger vein recognition algorithms developed based on other data sources to our data sets. As we know, there are significant differences between finger images from different sources due to different collection devices and corresponding light adjustment methods.
	Even if for two data sets collected by the same collection device, recognition algorithms may not perform consistently on them due to reasons including collection environments, user habits and collection time. Therefore, it is valuable to know whether algorithms developed based on other data sources perform well on our large-scale real-world data set. 
	
	Following FVRC 2016 competition, we set up three data sets, which are all invisible to candidates, with two for testing by candidates themselves before submitting their final algorithms and one for final evaluation after submission. ICFVR 2017 lasted from Feb 20 to May 20. 
This year, 11 teams from academic and industrial communities submitted their algorithms to our competition, 10 of which took part in the final evaluation. The information about them are displayed by Table \ref{table2} in the order of algorithm performance. In ICFVR 2017, 4 teams outperformed the winner of FVRC 2016 on the same database and with the same protocol. The winner of this year achieved 0.483\% EER far beyond 2.64\% achieved by the winner of last year. 
	
	
	\begin{table}[htbp]
		\begin{center}
			\begin{tabular}{|c|c|c|}
				\hline
				Team & Institution & No.\\
				\hline\hline
				W. Xu & \tabincell{c}{Shenzhen Maidi\\ Technology Co., LTD.} & T5\\
				\hline
				N.U. Ahmed & TigerIT & T2 \\
				\hline
				seacross & Seacross LTD. & T4\\
				\hline
				shakil & \tabincell{c}{Chittagong University of\\ Engineering \& Technology} & T3\\
				\hline
				YL. Jin & Peking University & T6\\
				\hline
				BIP & \tabincell{c}{South China University\\ of Technology} & T1\\
				\hline
				YJ. Chen~\etal & Peking University & T7\\
				\hline
				C. Hou~\etal & Peking University & T8\\
				\hline
				JX. Wen~\etal & Peking University & T9\\
				\hline
				ZY. Li~\etal & Peking University & T10\\
				\hline	
			\end{tabular}
		\end{center}
		\caption{Participating teams' names, institutions and corresponding No. Table is listed by participants' final rankings on the final evaluation data set. No. is arranged in order of registration time.}
		\label{table2}
	\end{table}

	\section{Benchmarks}
    \subsection{Strategies in RATE}
	
	RATE (Recognition Algorithm Test Engine) is an auto-rating system designed for finger vein recognition developed by Xian.\etal ~\cite{rate}. Unlike offline evaluation systems, which allow users to download data sets freely and upload evaluation results to compare with groundtruth such as Kaggle\footnote{https://www.kaggle.com/}, RATE is an independent-strongly supervised evaluation platform~\cite{cappelli2006performance}. We have strict constraints on input/output format, size of program and running time.
	
	Benchmark is a major part of RATE system. In brief, a benchmark consists of genuine pairs and imposter pairs produced by two different images. When generating a benchmark, RATE provides the following typical strategies:
		\begin{itemize}
			\item general. With sample number and class number set, it will generate genuine pairs by full permutation, followed by generating the same number of imposter pairs as genuine's.
			\item allInnerOneInter. For a view has $n$ classes and $s$ samples, it will generate $C_s^2 \times n$ genuine pairs\footnote{All possible combinations of $2$ elements out of $s$ elements times $n$}. A random sample is selected as the representative of a class. Then it will generate $C_n^2$ imposter pairs.
		\end{itemize}	
	
	\subsection{Data Sets and Benchmarks}
	
	The finger vein image used in RATE was collected from several real-world systems under different scenarios during Mar 2009 to Feb 2015. Every individual's index and middle fingers were collected. We guarantee that all individuals collected were aged under 30 and both male and female were involved sampled according to the ratio of men to women in China.
	
	The image format is bmp, 256 grayscale, 512*384 pixel resolution, captured by YANNAN Tech devices. Since all images are captured by the same type of device, samples from the same finger usually have the same size and orientation, but this is not guaranteed. 
	
    \begin{table}[htbp]
		\begin{center}
			\begin{tabular}{|c|c|c|c|c|}
				\hline
				Data Set & Size & Released & Visible & Difficulty \\
				\hline\hline
				DS0 & $50\times 5$ & Yes & Yes & Easy\\
				\hline 
				DS1 & $1000\times 5$ & Yes & No & Medium\\
				\hline 
				DS2 & $1000\times 5$ & Yes & No & Difficult\\
				\hline 
				DS3 & $1000\times 5$ & No & No & Hard\\
				\hline 
			\end{tabular}
		\end{center}
		\caption{General description of DS0,DS1,DS2,DS3}
		\label{table3}
	\end{table}
    
    \begin{table*}[htbp]	\begin{center}		\begin{tabu} to 0.7\linewidth {X[1,c,m] X[1,c,m] X[1,c,m] X[1,c,m] X[1,c,m]}		\toprule		Type & \multicolumn{2}{c}{Match} & \multicolumn{2}{c}{Non-Match} \\		\cmidrule(lr){2-3} \cmidrule(lr){4-5}				Data set & Image1 & Image2 & Image1 & Image2 \\		\midrule		DS1 & \includegraphics[width=0.13\textwidth]{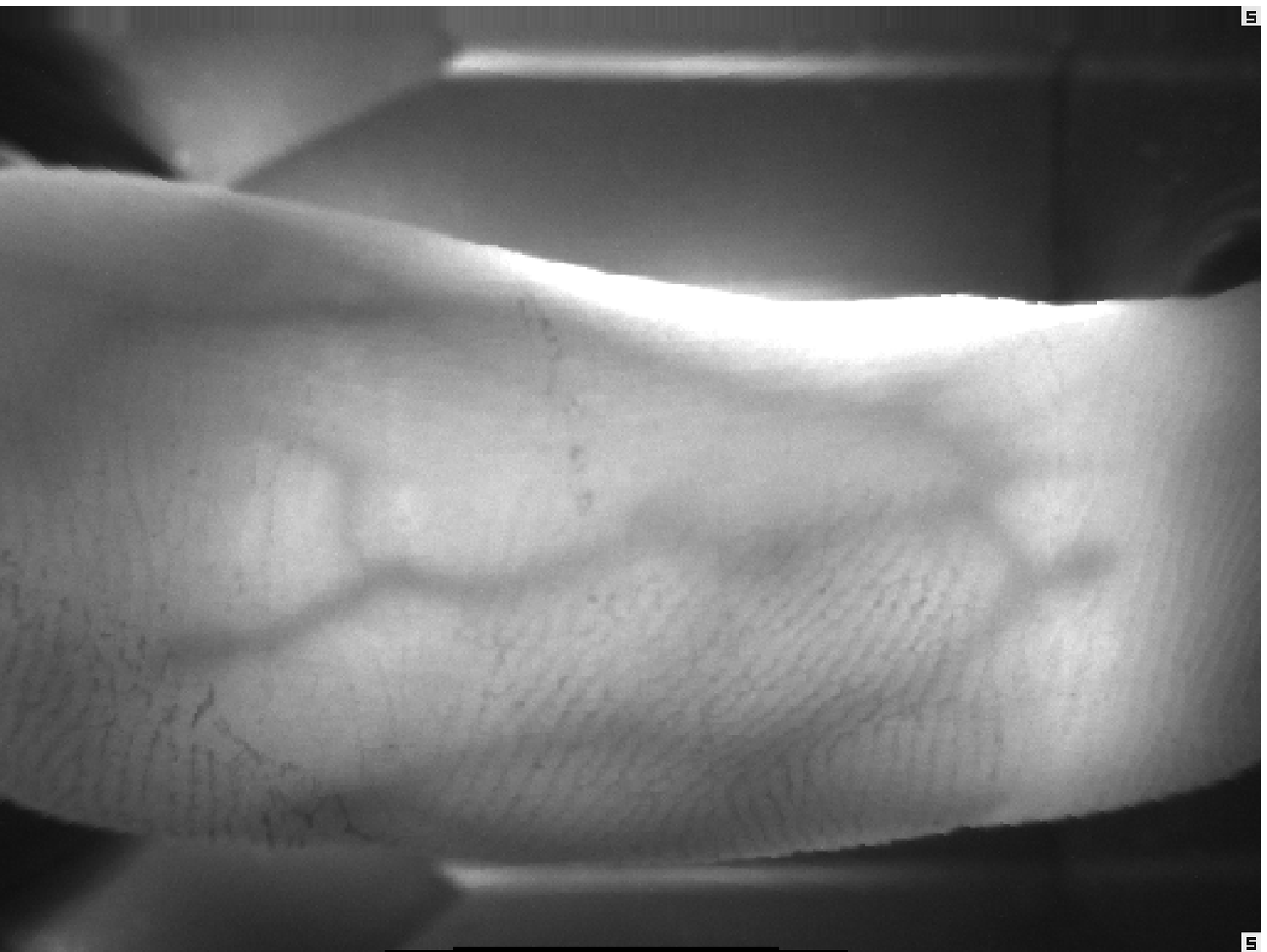} 		& \includegraphics[width=0.13\textwidth]{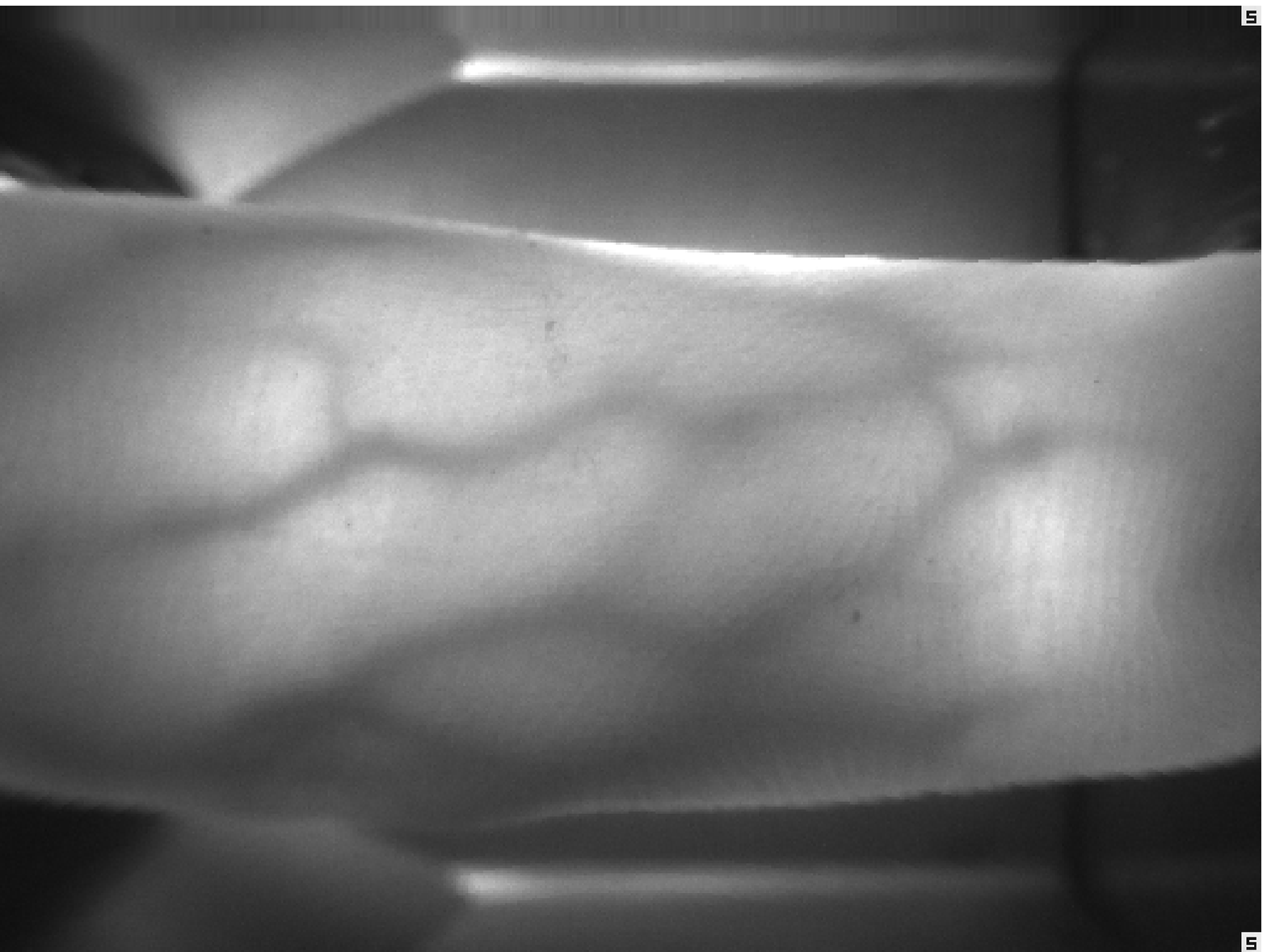} 		& \includegraphics[width=0.13\textwidth]{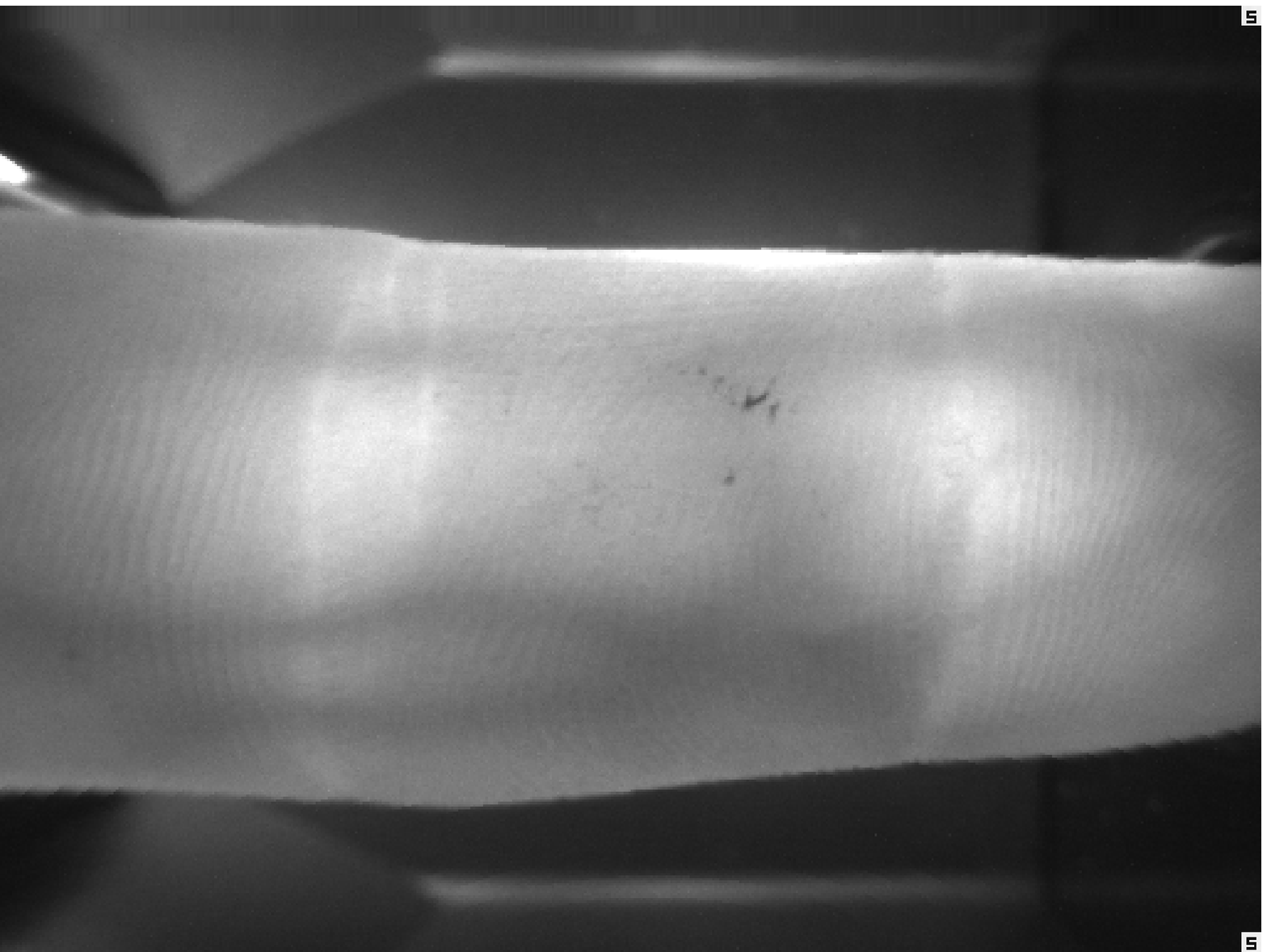} 		& \includegraphics[width=0.13\textwidth]{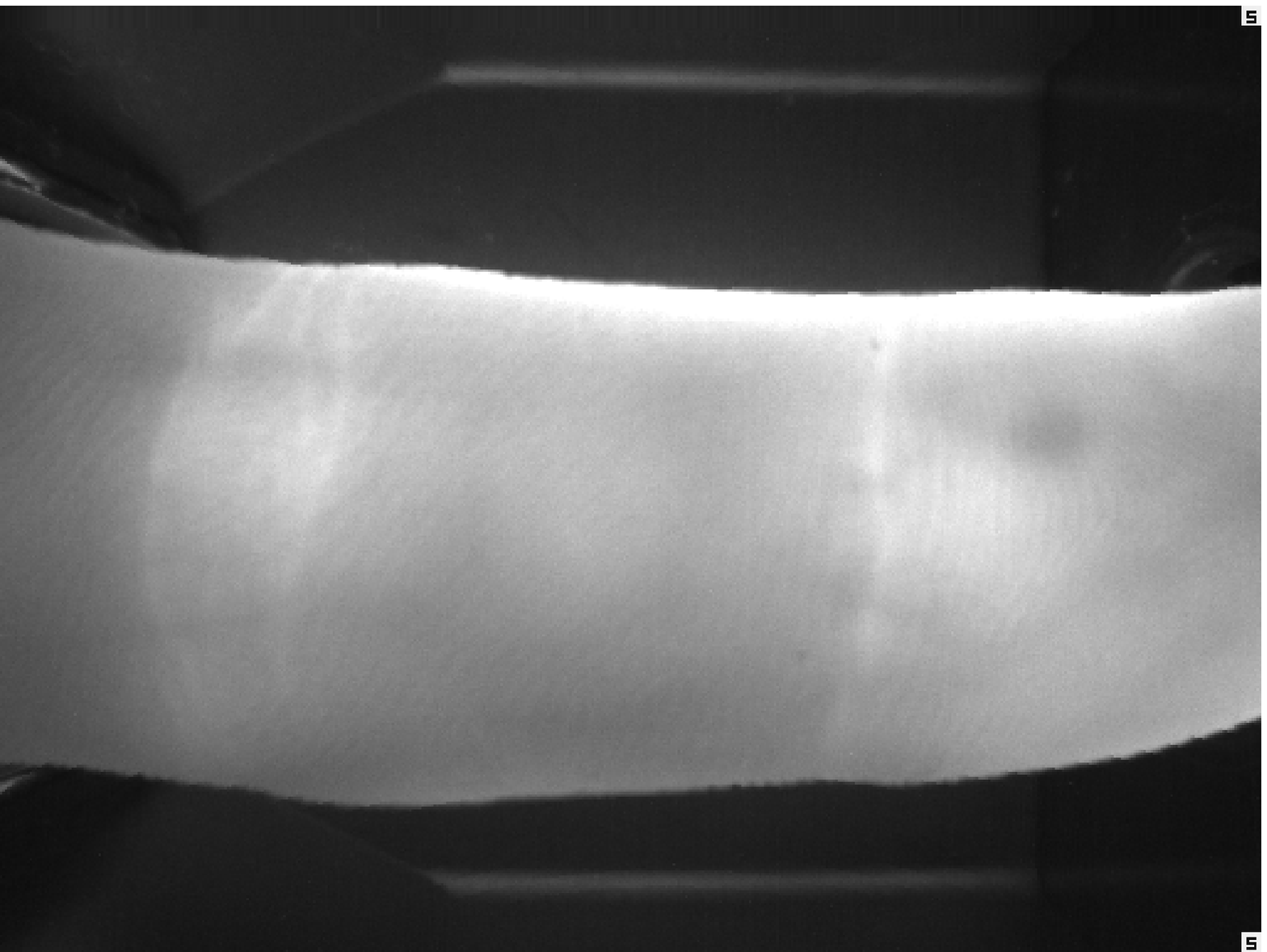} \\		\midrule		DS2 & \includegraphics[width=0.13\textwidth]{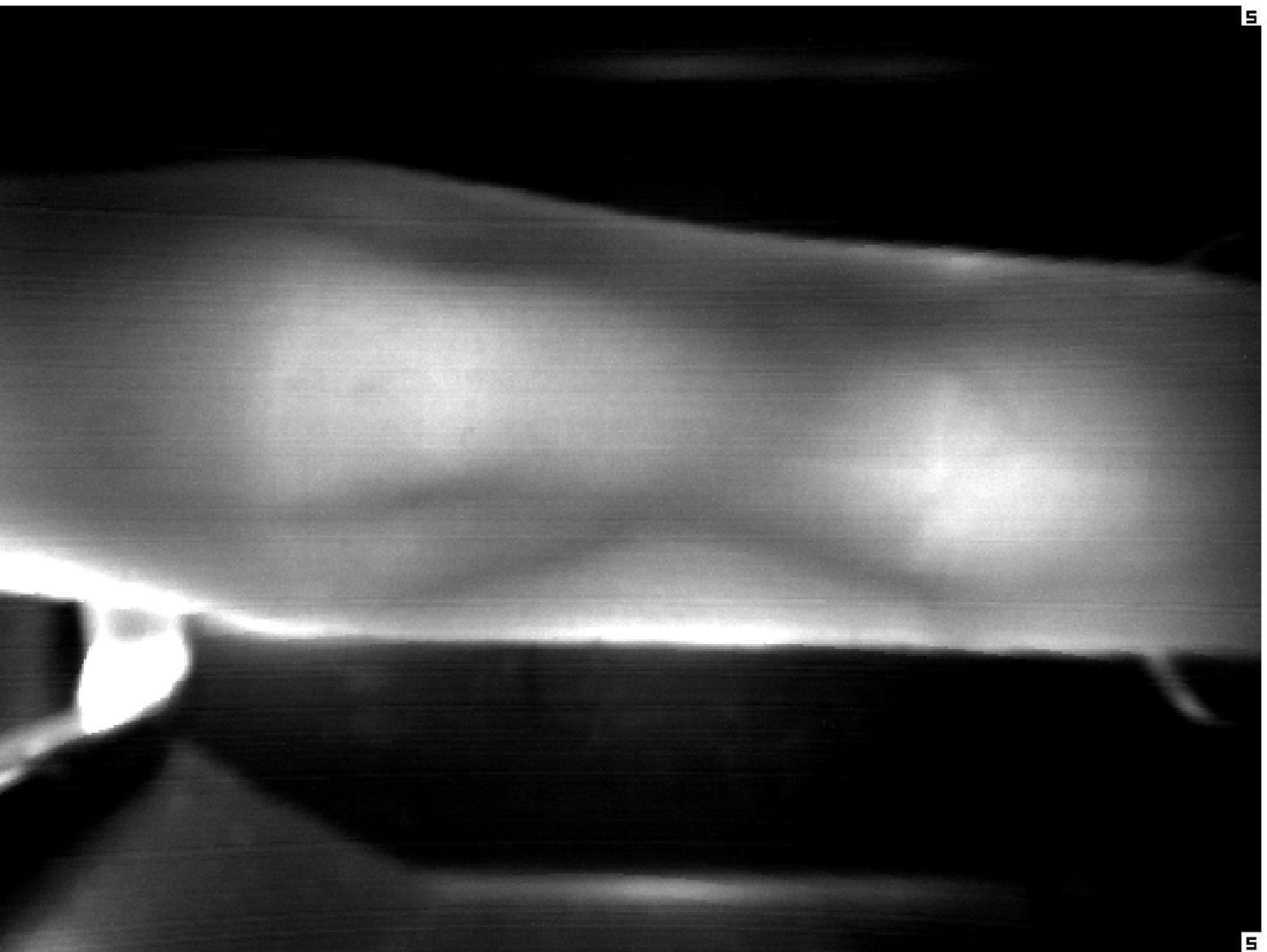} 		& \includegraphics[width=0.13\textwidth]{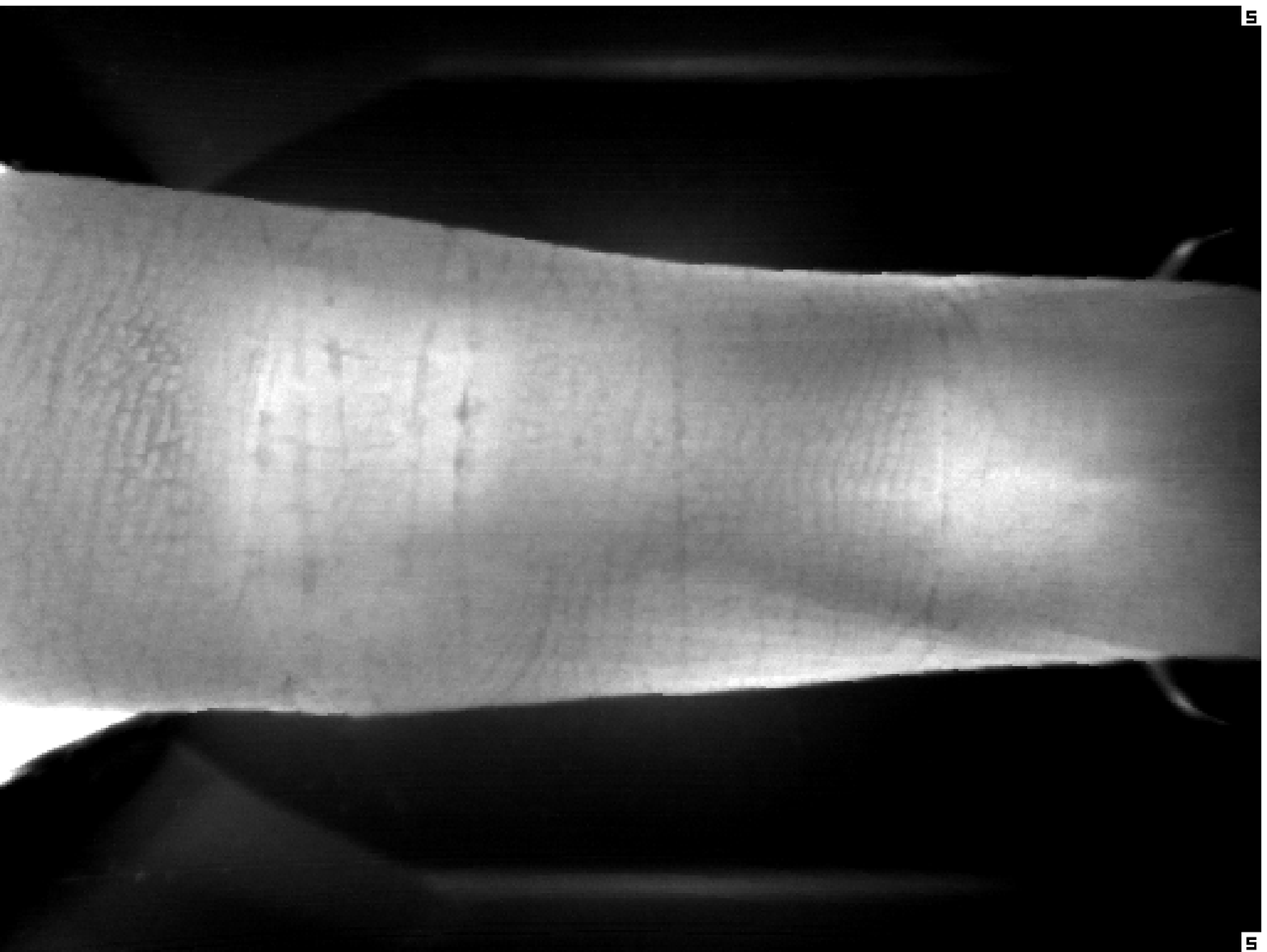} 		& \includegraphics[width=0.13\textwidth]{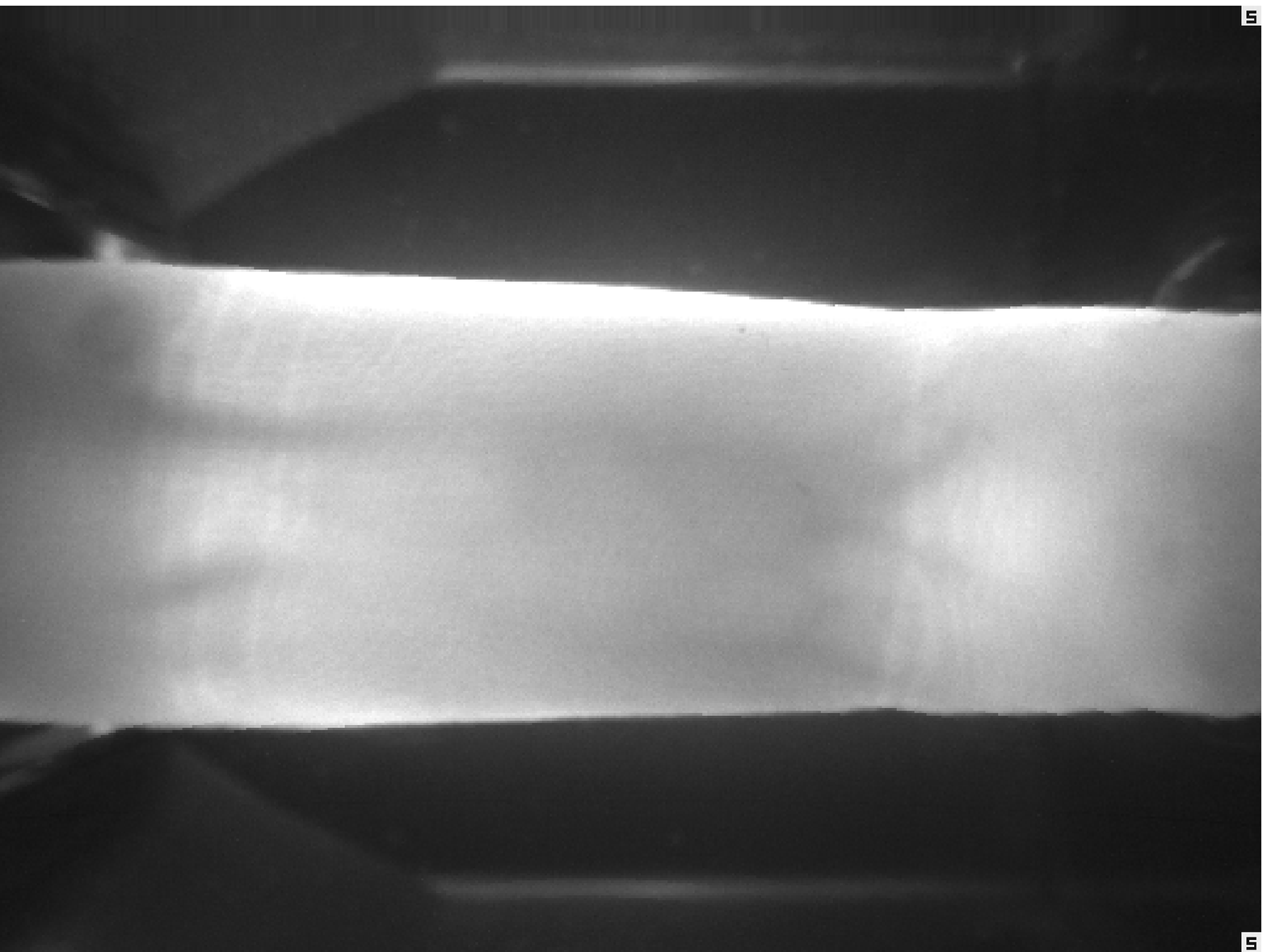} 		& \includegraphics[width=0.13\textwidth]{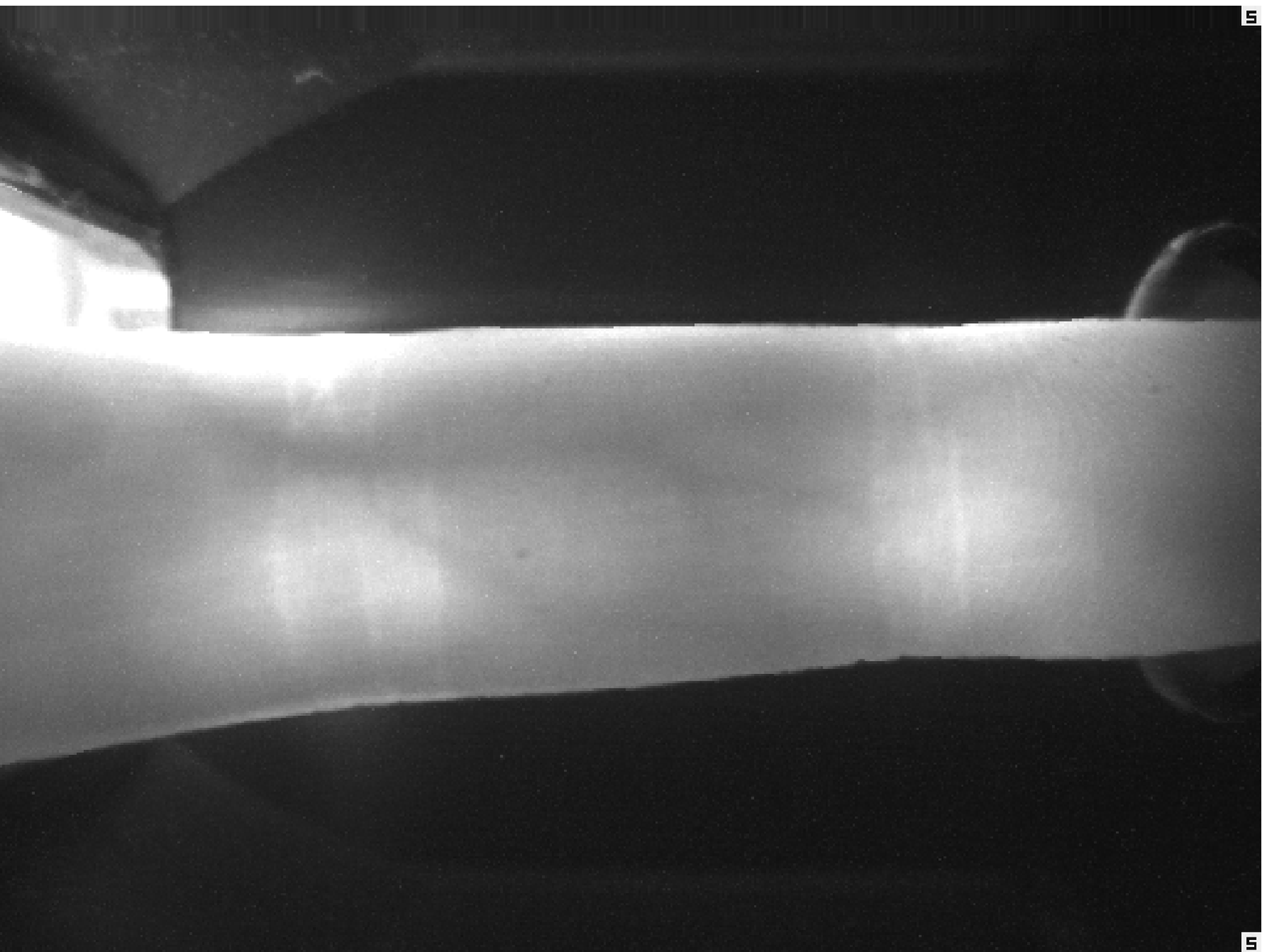} \\			\midrule			DS3 & \includegraphics[width=0.13\textwidth]{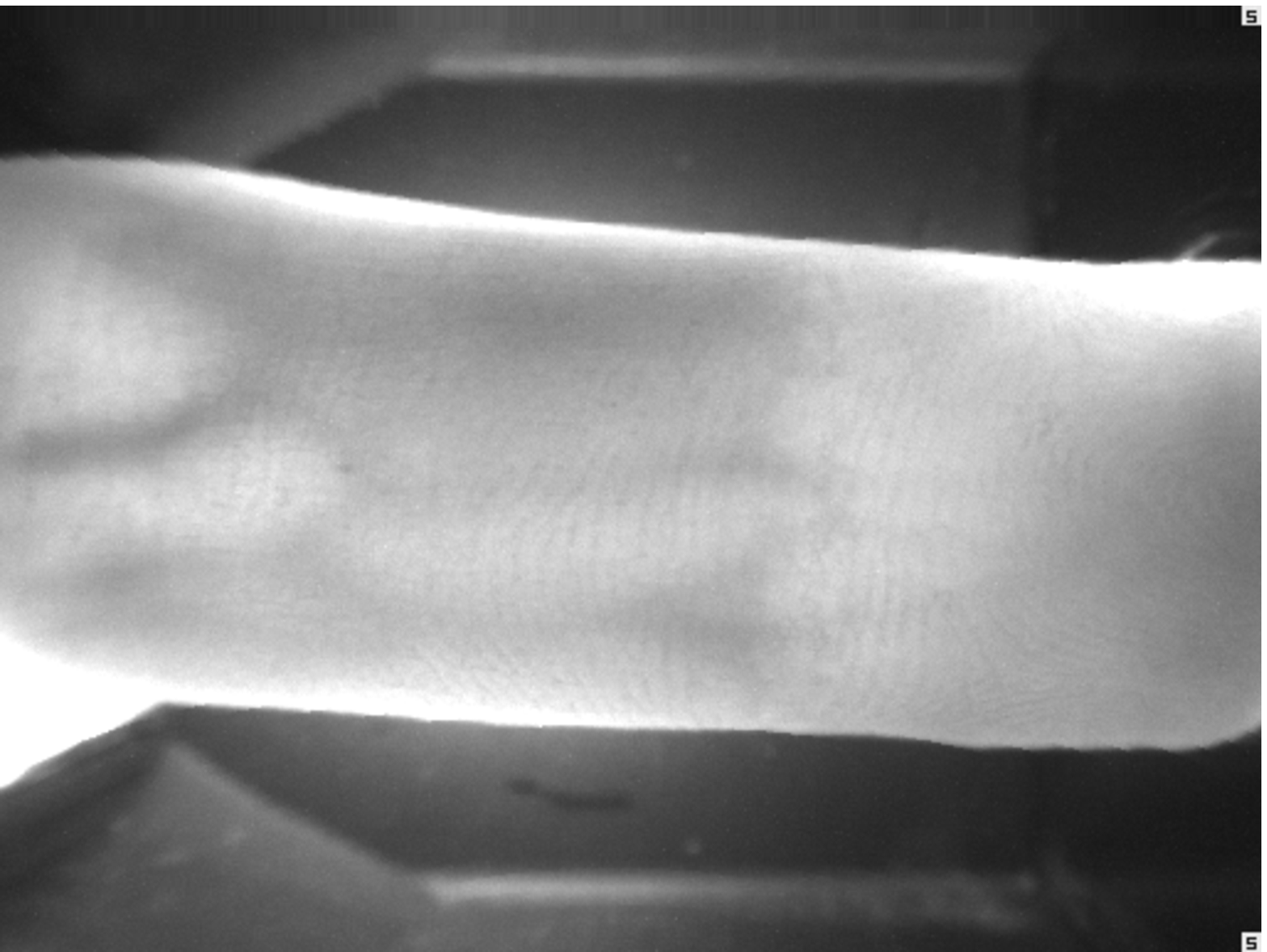} 		& \includegraphics[width=0.13\textwidth]{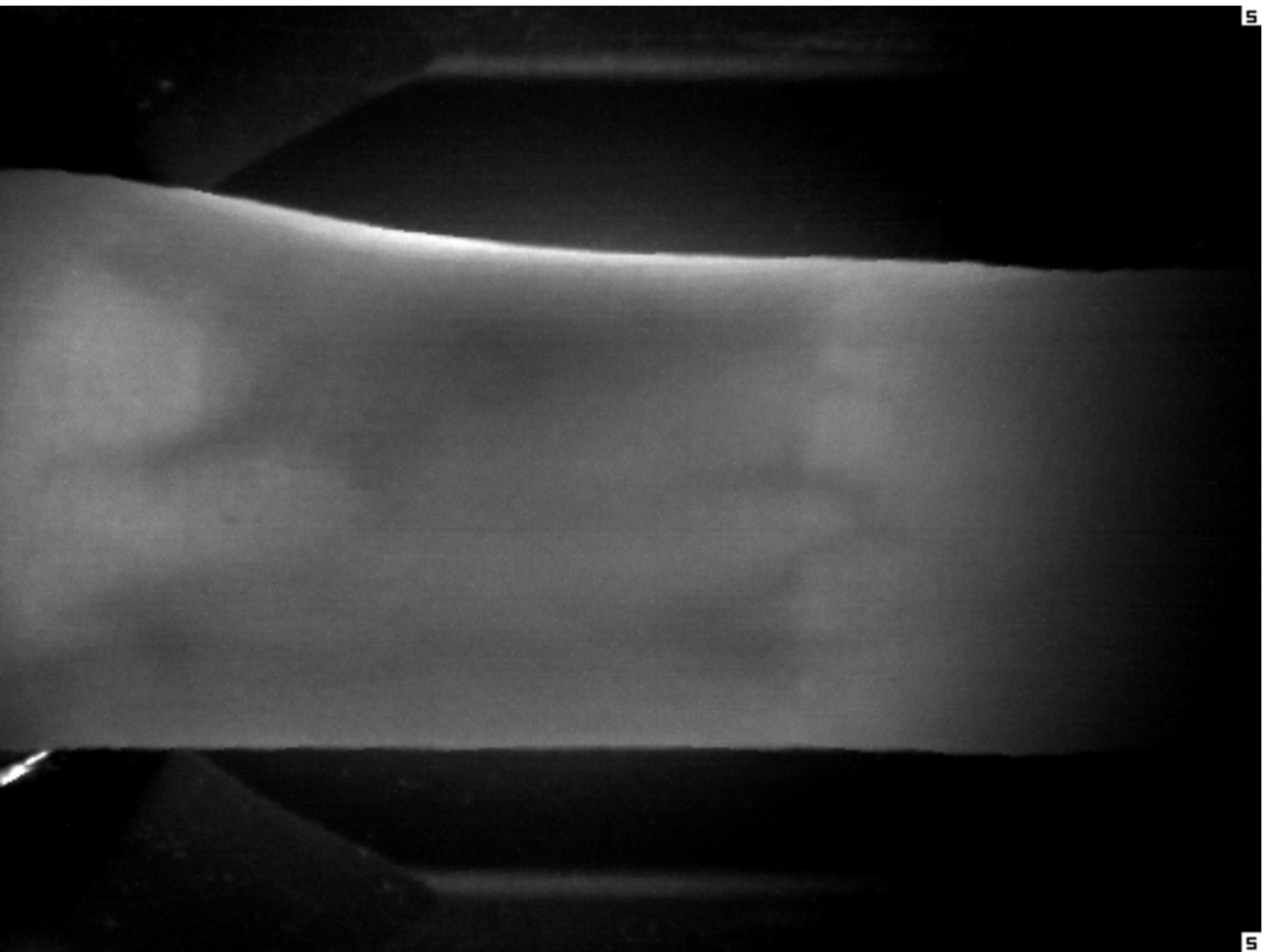} 		& \includegraphics[width=0.13\textwidth]{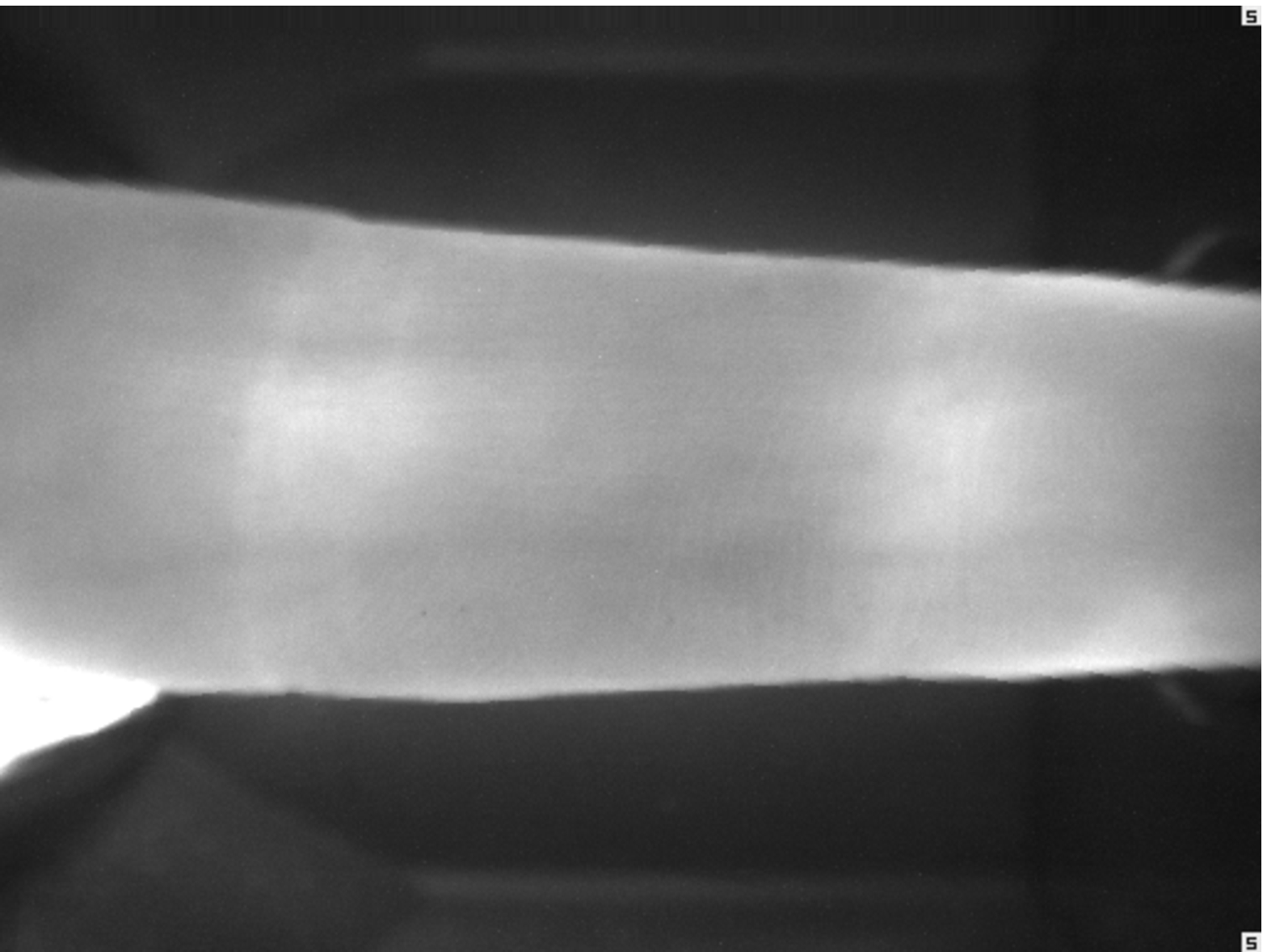} 		& \includegraphics[width=0.13\textwidth]{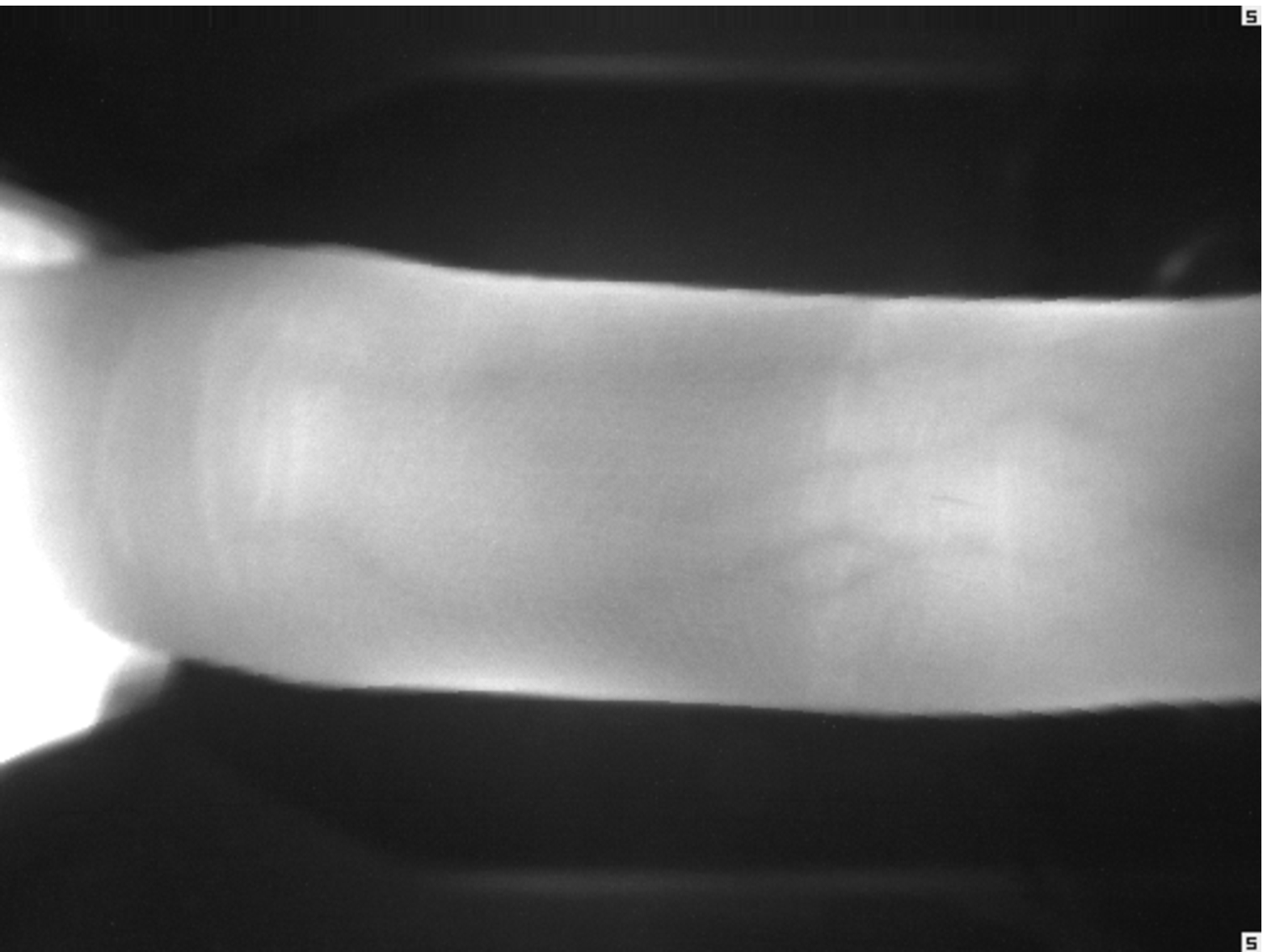} \\		\bottomrule		\end{tabu}	\end{center}
		\caption{Match and non-match pairs from DS1, DS2 and DS3.}  	\label{tab:samples} \end{table*}
    
	Four disjoint data sets were used in this competition, which are DS0, DS1, DS2, DS3. Table~\ref{table3} shows their general descriptions. A detailed description of their collecting scenarios is as follows.
    
	\begin{itemize}
		\item DS0. DS0 contains $50 \times 5$ samples including 50 fingers each with 5 samples. DS0 was collected in indoor environment and under guidance and supervision. This data set is for participants to test and debug their algorithms, which is suggested to be used as the starting to develop their algorithms.
		\item DS1. DS1 contains $1000\times 5$ samples, all of which were captured in indoor environment. The capturing process was under guidance and supervision. In general, the quality of this data set is relatively high, and competitors are expected to obtain good performances on this data set. The entire dataset was collected within one week, which is a relatively short time span.
		\item DS2. DS2 contains $1000\times 5$ samples, every subject of which was captured from real usage with a relatively longer time span (about 2 years) and wider age range compared to DS1, and was captured in a non-guidance outdoor environment. This data set is supposed to be more difficult to achieve as high performance as DS1. 
		\item DS3. DS3 contains $1000\times 5$ samples, which would be more difficult for the algorithm to verify than DS1 and DS2. Some of the data may show low similarity in one class, while some show high similarity between different classes.   
	\end{itemize}
	
	For that DS2 and DS3 were collected under non-guidance environment, some practical errors may exist, e.g. one finger class may contain other class's finger images. In order to eliminate this kind of mistakes, DS2 - DS3 were selected by a baseline algorithm first and then selected artificially to ensure that each class only contains its own finger images. The sample match and non-match pairs among DS1, DS2 and DS3 are shown in Table~\ref{tab:samples}. Compared with DS0 and DS1, DS2 and DS3 were collected under non-guidance environment. Thus DS2 and DS3 include more angles and gestures in general. 
	In short, the greater the time span, the greater the difficulty of data identification~\cite{kovacs2010new}.

	In the competition, DS0 was provided to competitors as raw images to help them understand our data set to debug. DS1 and DS2 generated two benchmarks separately. One benchmark was used for fast debugging using the above-mentioned general strategy that contains 10,000 genuine and imposter pairs. The other one was used to test the the trade-off of the algorithm between FMR and FNMR using allInnerOneInter strategy which contains 10,000 genuine pairs and 499,500 imposter pairs. DS1 and DS2 were released on RATE during the competition. Competitors could evaluate their algorithms with these data sets, which are de facto images invisible to them. DS3 generated a benchmark using the allInnerOneInter strategy for the final evaluation. DS3 was unreleased and competitors could not use or view DS3. The final ranking was evaluated according to the EER on the benchmark.

	\section{Algorithms}

	After the competition, we received the description of the algorithms from 6 teams that are listed in Table~\ref{table4} and Table~\ref{table5} by item. Detailed description of 4 algorithms were received in total, which are described in detail in the following text. Although we did not receive any algorithm relevant to machine learning, we still find some exciting and interesting results.
    
    \begin{table*}[htbp]
		\begin{center}
			\begin{tabular}{|c|c|c|c|c|c|c|}
				\hline
				& \multicolumn{6}{c|}{Enroll Part} \\ \cline{2-7} 
				& preprocess & \tabincell{c}{decrease or\\ remove the\\ effect of light} & resize & ROI & \tabincell{c}{rectify or \\adjust position \\and angle} & \tabincell{c}{used machine \\learning method} \\ 
				\hline\hline
				T5 & \tabincell{c}{Denoise \& \\ Smooth} & \tabincell{c}{using log \\ conversion} & $\checkmark$ & $\checkmark$ & $\checkmark$ & $\times$ \\ \hline
				T2 & Smooth & $\times$ & $\checkmark$ & $\checkmark$ & $\checkmark$ & $\times$ \\ \hline
				T3 & $\times$ & $\times$ & $\checkmark$ & $\checkmark$  & $\checkmark$ & $\times$ \\ \hline
				T6 & $\times$ & using filter & $\checkmark$ & $\checkmark$ & $\checkmark$ & $\times$ \\ \hline
				T7 & Denoise & $\times$ & $\checkmark$ & $\checkmark$ & $\times$ & $\times$ \\ \hline
				T9 & $\times$ & $\times$ & $\times$ & $\checkmark$ & $\times$ & $\times$ \\ \hline
			\end{tabular}
		\end{center}
		\caption{General description of algorithms' enroll part}
		\label{table4}
	\end{table*}
	
	\begin{table*}[htbp]
		\begin{center}
			\begin{tabular}{|c|c|c|c|}
				\hline
				& \multicolumn{3}{c|}{Match Part}  \\ \cline{2-4} 
				& \tabincell{c}{matching\\ method} & \tabincell{c}{used machine\\ learning method} & normalization \\ 
				\hline\hline
				T5 & Pixel Matching & $\times$ & $\checkmark$ \\ \hline
				T2 & Point Matching & $\times$ & $\checkmark$   \\ \hline
				T3 & Pixel Matching & $\times$ & $\times$   \\ \hline
				T6 & Pixel Matching & $\times$ & $\times$   \\ \hline
				T7 & Network Flow & $\times$ & $\checkmark$   \\ \hline
				T9 & Pixel Matching & $\times$ & $\times$ \\ \hline
			\end{tabular}
		\end{center}
		\caption{General description of algorithms' match part}
		\label{table5}
	\end{table*}
	
	\paragraph{T5} The algorithms consist of several steps including finger's edge detection, image pre-processing, feature extraction and feature comparision. The valid area of vein image can be obtained through edge detection. In the image pre-processing procedure, image filtering, area histogram equilization, curvature image fusion, etc. are used to enhance the vein information. In the feature extraction stage,  information extracted with good quality suggests that the vein feature is obtained successfully. In the feature comparision stage, based on the advanced information along with efficient matching search algorithm, the similarity between objects is calculated.
    
	\paragraph{T6} The Enroll section of the algorithm requires reading in the picture, simple
	edge detection, basic normalization to the posture of the finger and binary processing.
	
	The edge detection procedure of Enroll involves calling to the Canny function in OpenCV library, which uses gradient from a certain pixel to its neighboring pixels to recognize the boundaries of the picture. To eliminate outlier influences caused by noises or light, the algorithm implements a basic filter to
	the boundaries. With the observation that the boundaries of the finger must be continuous, the algorithm computes the average deviation in height between the current boundary point and its neighboring boundaries. When the average deviation exceeds a certain threshold, it will not be considered as contour of the finger.
	
	The normalization includes rotating the figure according to the middle line, 
	separating the finger restricted by the afore-mentioned contour. The middle line of the finger is acquired using the least squared regression of the middle points of the contour.
	The binary process uses the Wide Line Detector~\cite{huang2010finger}.
	
	The Match process involves moving the two templates by a deviation vector (x, y) and match template1[i+x][j+y] with template2[i][j] and get the maximum matched pixels. The maximum matched value divided by the total pixels will be output indicating the level of similarity.
	
	\paragraph{T7} 
	After putting the LBP image processing, Sobel operator edge detection, Sift feature extraction algorithm, etc., to trial to generate Enroll template in experiments, it is found that the wide line detector algorithm on the basis of isotropic nonlinear filtering mentioned in \cite{liu2007detecting} get the most suitable template for this competition. Through continuous adjustment of threshold in the implementation, the best threshold so far is obtained and the images in bmp format are converted to binary images.
	
	For that the processed images are sensitive to noise, the candidate uses the breadth first search algorithm to find and handle the connected area that is smaller than the threshold, which is regarded as vein area, so as to eliminate the impact of noise as much as possible.
	
	The author of matching algorithm chooses the construction of minimum cost and maximum flow algorithm for network flow, calculates the minimum cost and normalizes the result to be within 0 and 1 range. The author also adopts the sliding window method to find the largest matching value through sliding two images on a small scale, normalizes the largest matching value and outputs it.
	
	\paragraph{T9} The enroll step is to intercept an ROI area of the input image and save the grey scale information of the ROI area as template. The match step is to compare the grey scale information of the two templates. If the disparity of grey value for a pixel is less than 20, then we regard the pixel as matching successfully.
	
	The output value is the ratio of number of matching pixels to the total number of pixels.
	
	\section{Results and Analysis}
	\subsection{Protocol and Metrics}
	
	A submission consists of two Win32 executables, enroll.exe and match.exe, for the enrollment step and the matching step. The two executables should read input from the command line arguments and output results to the standard output or a designated file with the right exit value. They must obey the rules given in Table~\ref{table6}.
	
	\begin{table}[htbp]
		\begin{center}
			\begin{tabular}{|c|c|c|}
				\hline
				& Enroll.exe & Match.exe \\ \hline\hline
				Description & Create template & \tabincell{c}{Compare two \\templates} \\ \hline
				Input & \tabincell{c}{Paths of input \\ images and\\ output template file} & \tabincell{c}{Path of two \\templates for \\ comparisons} \\ \hline
				Output & None & \tabincell{c}{Similarity score \\ in {[}0-1.0{]}} \\ \hline
				Exit Code & \multicolumn{2}{c|}{0:success; else: fail} \\ \hline
				Time Limit & 30s & 10s \\ \hline
				Mem Limit & \multicolumn{2}{c|}{2048M} \\ \hline
				Size Limit & 20M & 300M \\ \hline
			\end{tabular}
		\end{center}
		\caption{Protocol for Enroll.exe and Match.exe}
		\label{table6}
	\end{table}
	
	We provide an Intel E5-2620 CPU, 32.0 GB memory and 1T 7200 rpm disk windows server for parallel evaluation. The queues of the evaluation algorithms are parallel, so each job will be treated fairly. On the basis of the previous two competitions, we relaxed the run time and memory limitations, providing more convenience for using other libraries and languages such as python and so on. On ICFVR~2017, participants submitted 168 versions of the algorithm, with at least 208 evaluations (excluding deleted ones) on DS1, DS2, and DS3. More than 35 million comparisons were computed by RATE in ICFVR~2017, compared to 10 million in FVRC~2015 and 34 million in FVRC~2016, respectively.
	
	As with FVRC~2016, the final ranking is only determined by the EER on a benchmark generated from DS3, using the allInnerOneInter strategy. For each task evaluated on the benchmarks of DS1, DS2 and DS3, RATE automatically generates some visible metrics for the participants to debug and analyze their own algorithms:
	\begin{itemize}
		\item False Match Rate (FMR) and False None-Match Rate (FNMR)
		\item Equal Error Rate (EER)
		\item FMR100 and FMR1000 (the lowest values of FNMR for FMR $\leq 1/100$ and $1/1000$, respectively)
		\item zeroFNMR (the lowest value of FMR for no False Non-Matches) and zeroFMR (the lowest value of FNMR for no False Matches)
		\item Fail to Enroll (FTE) and Fail to match (FTM)
		\item Average enrollment (Avg.E.T) time and average match time (Avg.M.T)
		\item Detection Error Trade-off (DET) curves
		\item Genuine and imposter score distribution histograms
	\end{itemize}
	
	\subsection{Competition Results}
	
\begin{figure}[htbp]
		\begin{center}
			\includegraphics[width=0.7\linewidth]{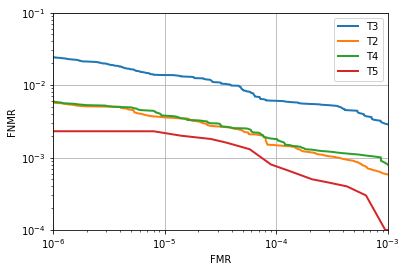}
		\end{center}
		\caption{Top 4 algorithms' DET on DS1}
		\label{fig:roc1}
	\end{figure}

\begin{figure}[htbp]
		\begin{center}
			\includegraphics[width=0.7\linewidth]{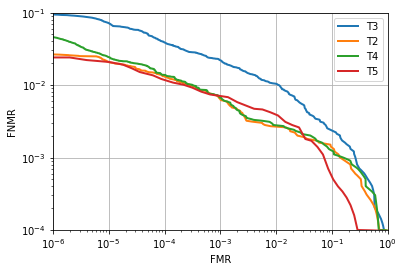}
		\end{center}
		\caption{Top 4 algorithms' DET on DS2}
		\label{fig:roc2}
	\end{figure}

\begin{figure}[htbp]
		\begin{center}
			\includegraphics[width=0.7\linewidth]{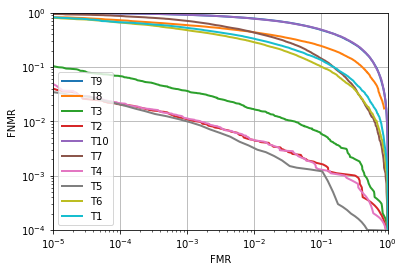}
		\end{center}
		\caption{DET on DS3}
		\label{fig:roc3}
	\end{figure}

	The detailed test results on DS3 of the algorithms is listed in Table~\ref{table9} and sorted by their final ranking. The DET curve trends of these algorithms are displayed in Figure~\ref{fig:roc3}. Table~\ref{table7} and Table~\ref{table8} show the evaluation results on DS1 and DS2 of top 4 algorithms among those that turn out superior to FVRC 2016. Figure~\ref{fig:roc1} and Figure~\ref{fig:roc2} display the DET curve of the afore-mentioned top 4 algorithms on DS1 and DS2.
	
	The results of DS3 show that the performances of algorithms of the competitors vary significantly. The top 4 algorithms have stable performances and better test results on DS1, DS2 and DS3 than that of FVRC 2016. The obvious performance degradation from DS1 to DS3 can be observed, which accords with our pre-set levels of difficulty.
	
	\begin{table}[htbp]
		\begin{center}
			\begin{tabular}{|c|c|c|c|c|c|c|c|}
				\hline
				& \rotatebox{90}{EER(\%)} & \rotatebox{90}{FMR100(\%)} & \rotatebox{90}{FMR1000(\%)} & \rotatebox{90}{zeroFNMR(\%)} & \rotatebox{90}{zeroFMR(\%)} & \rotatebox{90}{\#FTE} & \rotatebox{90}{\#FTM} \\ \hline\hline
				T5 & \textbf{0.035} & 0.01 & 0.01 & 0.1 & 0.23 & 0 & 58 \\ \hline
				T2 & 0.063 & 0.02 & 0.06 & 1.85 & 0.51 & 6 & 0 \\ \hline
				T4 & 0.084 & 0.04 & 0.08 & 1.74 & 0.53 & 5 & 2 \\ \hline
				T3 & 0.211 & 0.1 & 0.29 &  15.93 & 2.12 & 11 & 2\\ \hline
				*\cite{ye2016fvrc2016} & 0.77 & 0.73 & 1.41 & N/A & 7.44 & 0 & 0 \\ \hline
			\end{tabular}
		\end{center}
		\caption{Results on DS1. * is the best performance of FVRC~2016 on DS1.}
		\label{table7}
	\end{table}
	
	\begin{table}[htbp]
		\begin{center}
			\begin{tabular}{|c|c|c|c|c|c|c|c|}
				\hline
				& \rotatebox{90}{EER(\%)} & \rotatebox{90}{FMR100(\%)} & \rotatebox{90}{FMR1000(\%)} & \rotatebox{90}{zeroFNMR(\%)} & \rotatebox{90}{zeroFMR(\%)} & \rotatebox{90}{\#FTE} & \rotatebox{90}{\#FTM} \\ \hline\hline
				T5 & 0.437 & 0.38 & 0.68 & 28.82 & 2.21 & 0 & 63 \\ \hline
				T2 & \textbf{0.311} & 0.27 & 0.64 & 70.71 & 2.51 & 5 & 183305 \\ \hline
				T4 & 0.341 & 0.28 & 0.67 & 69.19 & 3.06 & 10 & 213602 \\ \hline
				T3 & 1.027 & 1.07 & 2.21 &  85.62 & 9.17 & 14 & 20\\ \hline
				* & 1.8 & 1.97 & 3.7 & N/A & 13.6 & 0 & 0 \\ \hline
			\end{tabular}
		\end{center}
		\caption{Results on DS2. * is the best performance of FVRC~2016 on DS2.}
		\label{table8}
	\end{table}
	
	\begin{table*}[htbp]
		\begin{center}
			\begin{tabular}{|c|c|c|c|c|c|c|c|c|c|c|}
				\hline
				& \rotatebox{90}{EER(\%)} & \rotatebox{90}{FMR100(\%)} & \rotatebox{90}{FMR1000(\%)} & \rotatebox{90}{zeroFNMR(\%)} & \rotatebox{90}{zeroFMR(\%)} & \rotatebox{90}{\#FTE} & \rotatebox{90}{\#FTM} &
				\rotatebox{90}{Avg.E.T(ms)} &
				\rotatebox{90}{Avg.M.T(ms)} &
				\rotatebox{90}{Avg.Template Size(KB)}
				\\ \hline\hline
				T5 & \textbf{0.483} & 0.3 & 0.96 & 49.74 & 3.98 & 0 & 0 & 64.33 & 5.28 & 13 \\ \hline
				T2 & 0.581 & 0.46 & 1.09 & 93.99 & 4.54 & 4 & 0 & 39.9 & 20.09 & 81 \\ \hline
				T4 & 0.593 & 0.46 & 1.14 & 94.03 & 6.31 & 12 & 2 & 60.3 & 24.85 & 81 \\ \hline
				T3 & 1.498 & 1.67 & 3.62 &  98.29 & 15.34 & 13 & 21 & 20.64 & 14.39 & 57 \\ \hline
				T6 & 10.12 & 28.6 & 47.29 &  99.98 & 84.14 & 0 & 0 & 8.04 & 11.75 & 12 \\ \hline
				T1 & 11.91 & 33.16 & 52.28 & 100.0 & 82.62 & 0 & 0 & 7.07 & 6.08 & 8 \\ \hline
				T7 & 12.37 & 43.15 &70.27  & 98.53 & 96.55 & 0 & 2 & 8.48 & 7.64 & 12 \\ \hline
				T8 & 18.51 & 43.44 & 58.73 & 100.0 & 85.44 & 0 & 5661 & 25.87 & 18.03 & 13 \\ \hline
				T9 & 28.46 & 77.69 & 90.96 & 100.0 & 99.05 & 0 & 0 & 22.64 & 15.06 & 115 \\ \hline
				T10 &28.92 & 77.29 & 91.06 & 100.0 & 99.35 & 0 & 16 & 18.37 & 11.16 & 101 \\ \hline
				*Best of FVRC~2016~\cite{ye2016fvrc2016} & 2.64 & 3.29 & 5.86 & 100.0 & 20.71 & 0 & 0 & 16.91 & 9.57 & 20\\ \hline
			\end{tabular}
		\end{center}
		\caption{Results on DS3}
		\label{table9}
	\end{table*}
	
	\subsection{Analysis}
	
	Based on results in Section 5.2, We analyze the adaptiveness to acquisition condition, impact of failure, and efficiency of the algorithms.
	
	In Section 3, we introduce the collecting conditions of each data set. The test results imply that the collecting environment has a significant influence on test results. An intricate environment increases the recognition difficulty. The main factors affecting the results are light and the finger gesture. The finger images acquired indoor are with good lighting, while the images acquired outdoors are prone to light overexposure, light leak and light underexposure, which increases the difficulty of extracting features of veins. Whether to provide instructions and supervision or not actually determines the normalization of gesture. The inappropriate poses of the finger stretching into the acquisition device can lead to panning, rotation, tortuosity, etc, which increases the difficulty of feature extraction and feature matching. Meanwhile, the range of time is an influential factor which is also intricate for the following facts. For one thing, it should be noted that the images acquired in different seasons and under different temperatures vary in condition. For that, in a short period of time, the fluctuation of body condition and vein is minor, while the long range of time increase the uncertainty of the fluctuation and brings challenges for recognition. For another thing, long time of usage cultivate users' correct usage.
	
	The feedback shows that better performed algorithms adopt some strategies to abate the afore-mentioned impact of noise. The T5 algorithm which ranks 1st reduces or removes the impact of lighting, and rectifies the position and angle of the fingers. T2 algorithm which ranks 2nd and T4 algorithm which rank 3rd rectify the position and angle of the fingers while not take the impact of light into consideration. The T6 algorithm which ranks 5th uses a filter to abate the impact of noise and light. The T7 algorithm uses breadth first search algorithm, finds the connected area which is less than the threshold and eliminates the impact of noise as much as possible. T9 algorithm, which ranks the 9th, adopts no measure to eliminate noise or rectify the angle and position.
    
	Although on DS3, there exists some degree of failure to enroll and match for 5 algorithms and T8 algorithm has a large ratio of FTM which is up to 1.11\%. On a strongly supervised evaluation platform, there may be many factors that lead to failure, such as timeout, crash, template limit and missing template~\etal. As a consequence, algorithms rejecting poor quality fingerprints at enrollment time could be implicitly favored since many problematic comparisons could be avoided~\cite{cappelli2006performance}. Based on this, we observe an interesting phenomenon that the best-performing algorithm T5 still ranks first on DS1, but only ranks 3rd on DS2, which distinguishes from the best algorithm in the last year. On DS2, the top 2 algorithms have a lot of failures in matching. We think that the reason may lie in our methods for calculating FMR and FNMR, in which we do not take attempt pairs of failures into consideration. We consider using the following formula to revise the calculation of FMR and FNMR. 
    \begin{align*}
	\text{FMR} &= \frac{\text{succeed imposter pairs} - \sum I(score \leq t) }{\text{imposter pairs}} \\
	\text{FNMR} &= 1 - \frac{\text{succeed genuine pairs} - \sum I(score \leq t) }{\text{genuine pairs}}
	\end{align*}
	where $t$ is a pre-set threshold, and $I$ is an indicator function, which indicates the number of pairs whose scores are not greater than threshold t.
	
	Besides EER of the algorithms, the efficiency is an important aspect for us to inspect. In ICFVR~2017, the restriction for memory is relatively relaxed. Thus, there is no statistical survey for memory usage. However, the time consumed and size of the template are of our concern. Compared to T2, T4 and T3 algorithms, the size of the template extracted from T5 algorithm is smaller. Compared to all algorithms, T5 algorithm consumes most time in the enroll part, but consumes least time in match part, which indicates that the template extracted from T5 presents better expressiveness for features. The template extracted by T9 algorithm is the largest, which is in accordance with their description--simply intercept ROI and output pixel directly.
	
	\section{Conclusion}
	
	In order to explore the performance of more excellent algorithms in our data sets, we successfully hosted ICFVR~2017 on the basis of two previous competitions. We adopted the three data sets used by FVRC~2016, which were captured in different scenarios. In the end, we received 10 algorithms submitted by teams from industry and academia, with four teams outperforming FVRC~2016's best results. Moreover, we received description of the details about their algorithms by six teams, which helped us to analyze the results in detail.
	
	In addition to the research of the algorithm itself, the algorithms and descriptions collected in competition can also be used for the relevant research, such as algorithm fusion. In subsequent competitions, we are to introduce larger-scale data sets, as well as attract more algorithms including machine learning algorithms.

	{\small
		\bibliographystyle{ieee}
		\bibliography{icfvr2017}
	}
	
\end{document}